%% file: aaai23.tex
\documentclass[letterpaper]{article} 
\usepackage{aaai23}  
\usepackage{times}  
\usepackage{helvet}  
\usepackage{courier}  
\usepackage[hyphens]{url}  
\usepackage{graphicx} 
\usepackage{natbib}  
\usepackage{caption} 
\usepackage{algorithm}

\usepackage{newfloat}
\usepackage{listings}
\DeclareCaptionStyle{ruled}{labelfont=normalfont,labelsep=colon,strut=off} 
\floatstyle{ruled}
\newfloat{listing}{tb}{lst}{}
\floatname{listing}{Listing}

\usepackage{amsfonts}
\usepackage{amsthm}
\usepackage{enumerate}
\usepackage{amsmath}
\usepackage{bbm}
\usepackage{algorithm}
\usepackage[noend]{algpseudocode}
\algnewcommand\algorithmicinput{\textbf{Input:}}
\algnewcommand\algorithmicoutput{\textbf{Output:}}
\algnewcommand\Input{\item[\algorithmicinput]}%
\algnewcommand\Output{\item[\algorithmicoutput]}%
\usepackage{subfig}
\nocopyright

\title{Turning Mathematics Problems into Games: Reinforcement Learning \\ and Gr\"obner bases together solve Integer Feasibility Problems}

\author{
    Yue Wu\thanks{Equal contribution},
    Jes\'us A. De Loera$^*$
}
\affiliations {
    University of California, Davis\\
    yvwu@ucdavis.edu, deloera@math.ucdavis.edu
}

\input{dfn.tex}

\begin{document}

\maketitle

\begin{abstract} Can agents be trained to answer difficult mathematical questions by playing a game? We consider the \emph{integer feasibility problem}, a challenge of deciding whether a system of linear equations and inequalities has a solution with integer values. This is a famous NP-complete problem with applications in many areas of Mathematics and Computer Science. Our paper describes a novel algebraic reinforcement learning framework that allows an agent to play a game equivalent to the integer feasibility problem.
We explain how to transform the integer feasibility problem into a  game over a set of arrays with fixed margin sums. The game starts with an initial state (an array), and by applying a legal move that leaves the margins unchanged, we aim to eventually reach a winning state with zeros in specific positions. To win the game the player must find a path between the initial state and a final terminal winning state, if one exists. Finding such a winning state is equivalent to solving the integer feasibility problem. The key algebraic ingredient is a Gr\"obner basis of the toric ideal for the underlying axial transportation polyhedron. The Gr\"obner basis can be seen as a set of connecting moves (actions) of the game. We then propose a novel RL approach that trains an agent to predict moves in continuous space to cope with the large size of action space. The continuous move is then projected onto the set of legal moves so that the path always leads to valid states.  As a proof of concept we demonstrate in experiments that our agent can play well the simplest version of our game for 2-way tables. Our work highlights the potential to train agents to solve non-trivial mathematical queries through contemporary machine learning methods used to train agents to play games.
\end{abstract}

\section{Introduction}
Reinforcement learning has seen tremendous success in recent years, especially in playing games at levels that achieve superhuman performances \cite{human_level_control, Silveretal-Nature17,Guoetal-IJCAI16,Vinyalsetal-Nature19}). 
The philosophical principle we introduce in this paper is to try to reformulate non-trivial mathematical problems as games and then try to adapt reinforcement learning techniques to play those games. By winning the game we solve the original mathematical problem. Of course this first requires (at least for now) a human creating the right game for the given mathematical problem. As a proof of concept, in this paper we use the \emph{integer feasibility problem (IFP)}.
In its simplest form, the IFP is the decision question of whether
a polyhedral system $\{x : Ax=b, \ x \geq 0\}$ has an integer vector solution $x$ (note that any system of equations and inequalities can be reduced to this standard form). This famous NP-complete problem is very important in mathematical optimization, discrete mathematics, algebra, and other areas of mathematics. What we propose here is to turn it into a game on arrays. 

As we explain below we transform every instance of the integer IFP into a positional game over arrays or tables with fixed margin sums belonging 
to the axial transportation we use an old polynomial-time algorithm from \cite{DeLoeraOnn-04} that canonically rewrites an instance of the IFP as the face of a $l \times m \times n$ axial transportation polytope. Axial transportation polytopes are convex polytopes whose points are arrays or tables with fixed axial sums of entries \cite{YKK}. The second idea is that we know well the toric (polynomial ring) ideals of axial transportation polytopes have an explicit North-West Gr\"obner basis that connects all lattice points. 
The game starts with an initial state (an initial array), and by applying a legal move that leaves the margins unchanged, we aim to reach a winning state with zeros in specific positions. To win the game the player must find a path between the initial state and a final terminal winning state. Finding such a winning state is equivalent to solving the integer feasibility problem. As we see later the winning position, if one exists, is essentially a table with very specific zero entries.
The Gr\"obner basis of the toric ideal for the underlying transportation polyhedron as a set generating all connecting moves that the agent can be trained to select the moves. Of course our games sometimes have no solution, then our method, at least for now, may have to exhaust all positions before concluding this.
Reinforcement learning tools are then used to search the game space 

Reinforcement learning on games has achieved great success and a key point of our paper is that the success can be extended to train artificial agents to play these games with a mathematical origin.
We make a successful practical demonstration of these ideas with experiments in the simplest form of our games, the situation for 2D tables where we trained an agent to predict the moves. Our game can be viewed as a variant of the \emph{stochastic shortest path problem}~\cite{ssp,ssp_Bertsekas13} where an agent takes a path to reach pre-defined goal states. However, unlike any of the games like Go, maze, etc, our table game has a very large and highly (algebraically) structured action space. In 2-way tables, the minimum Grobner basis contains moves of degree $4$ ($4$ non-zero entries). Although these most basic moves are sufficient for solving the game, the progress they make at each step is limited, which can cause extremely long episodes. We instead consider much more advanced moves that are integer combinations of the minimum moves. These complicated moves will be very effective for solving the table games faster. New techniques are required to compute these advanced moves because they are not only very large in number but also difficult to construct. We treat it as a regression problem and make the agent predict continuous moves which are then projected onto its closest lattice point in a constraint set specified by an integer program. We make the whole framework differentiable and thus can be trained end-to-end. 
 
 We conduct experiments on 2-way table and the results are promising, which sheds light on the potential of \Grobner basis approaches for integer feasibility problems. 


\section{Basic notions: polyhedra and Gr\"obner bases} \label{prelim}

We begin describing how testing integer feasibility is actually 
equivalent to playing a certain game on finite set of arrays inside a polytope. 
We begin with some notation: Let $l$,$m$ and $n$ be three positive
integers. Let $x=(x_1,\dots,x_l)$, $y=(y_1,\dots,y_m)$, and
$z=(z_1,\dots,z_n)$ be three rational vectors of lengths $l$, $m$ and
$n$, respectively, with non-negative entries.  We consider the 3-way
transportation polytope $T_{x,y,z}$ with entries $a_{i,j,k}$ defined by 1-marginals, 
\begin{align*}
    & a_{i,j,k} \geq 0 ,\, \forall i,j,k,\quad
 \sum_{j,k} a_{i,j,k} = x_{i},\, \forall i \\
& \sum_{i,k} a_{i,j,k} = y_{j},\, \forall j,\quad
 \sum_{i,j} a_{i,j,k} = z_{k},\, \forall k.
\end{align*}

The \emph{margins} are sums of the certain entries of an 3-way array of numbers.
We begin with stating the first algorithmic step of this paper:

\begin{theorem} \label{theorem:main}
Any rational convex polyhedron can be written as a face $F$ of some $3$-way
transportation polytope $Q$ with $1$-marginals $x,y,z$ supported by a hyperplane
of the form $\sum_{(i,j,k) \in S} a_{i,j,k}=0$, where $S$ is a finite
subset of indices. The sizes $l,m,n$, the set $S$, and marginals $x,y,z$ 
can be computed explicitly in polynomial time on the size of the input.
Specifically, the face $F$ is given by certain entries forced to be zero.
\end{theorem}


This theorem is a particular case of a much more general theory further developed in \cite{DeLoeraOnn-04,DeLoeraOnn-SIAM04}. Given a convex polytope $P=\{x : Ax=b, \ x \geq 0\}$ (for which we wonder whether there is 
a lattice point), the very first step is to use Theorem \ref{theorem:main} to transfer  the points of $P$ as 3-way tables. We construct $1$-marginals for a $3$-way transportation polytope $Q$, and a set $S$ of triples $(i,j,k)$, such that $P$ is the face of $Q$ of those points with $a_{ijk}=0$ for all $i,j,k$ in $S$.

Unlike general polytopes, for axial transportation polytopes with given
1-marginals it is trivial to decide integer feasibility and to
find an integer vertex of $Q$. This can be done via the famous greedy
\emph{north-west-up corner} rule \cite{hoffman,YKK}. In fact, integral points 
of $Q$ will exists as long as the polytope is non-empty and the 1-marginals
are integral. To check if $Q$ is nonempty over the reals (i.e., LP feasibility), 
all we have to do is check whether the sum of the entries in the three
margins $x,y,z$ are equal. 

The tables inside $Q$ will be the state space for the game.  
Next, in order to detect whether we have an integral solution of $P$, 
we can use another property of axial transportation polytopes, namely 
we prove that there exists an explicit Gr\"obner basis for the toric 
ideal of axial transportation polytopes which will give an integer 
point in $P$ if one exists. As explained in
\citet{GB+CP} one can rely on the Gr\"obner
basis elements as moves, which move us from one
feasible integer point, a table, in $Q$ to another. We stress that 
although Gr\"obner bases have been criticized as having too 
many elements and potentially many elements of large
entries (hence being hard to compute), in our situation we
have nicer structure that aids to practical performance. 
Unlike prior situations our Gr\"obner bases can be generated on 
the fly and their elements have entries only $0,1,-1$ due to the special structure of axial transportation polytopes. We move from table to table that satisfy 
same margins. In other words, we get around the
nasty structure of an arbitrary polyhedron $P$ via the canonical embedding
$P$ as a face of an axial transportation polyhedron (which is a larger polyhedron, but it is simpler).  By our results in Section 
\ref{algorithm} the pivots will find a feasible point in $P$ if and only if one exists. 

Let us explain next more about Gr\"obner bases and the way to always find an initial table. While general transportation polytopes can be as nasty as arbitrary polytopes, axial transportation polytopes have several noteworthy computational advantages.
First, checking real feasibility is trivial, the polytope is feasible
if and only if the sum of the entries in each of the 1-margins equal
the same value. More strongly, there will be an \emph{integer}
solution if the margins are integer. 

A special case of axial transportation polytopes that is familiar to most readers, consists of 2-way transportation polytopes. They appear in most college-level optimization courses as bipartite network-flow problems. The $n \times m$ Hitchcock transportation problem: $\min_x c[i,j]x_{ij}$, s.t. $\sum_{i=1}^n x_{ij} =a^1_{j} \ \sum_{j=1}^n x_{ij}=a^2_{i}$.
An initial feasible solution can be obtained from a
greedy procedure for certain sequences of the variables, the
\emph{northwest-corner rule}. This algorithm proceeds along a sequence
of entries $S$ and maximizes each variable in turn with respect to the
margins that bound it. Its running time is $O(nm)$. In general the
algorithm constructs only a feasible, not an optimal solution.
Fortunately, \citet{hoffman} gave a sufficient and necessary condition on
$S$ such that the algorithm always constructs an optimal solution for
arbitrary demand and supply vectors $a^1$ and $a^2$ and cost vector
$c$. But for some cost vectors and special sequences the solution will
be optimal. Sequences and cost matrices which fulfill the property
above are called {\em Monge sequences}. 

For this paper, we need to find an initial integer table efficiently for $3$-way axial transportation polytopes. The good news is that, once again, a very 
similar greedy algorithm for the classical $2$-way problem explained above can be applied to obtain a feasible solution for this more general case. In pseudocode
this algorithm will read as follows in the case of a 3-way $l \times m \times n$ 
axial transportation polytope: Take an arbitrary order of the variables, say the sequence
$S:= ([i_1,j_1,k_1], [i_2,j_2,k_2] ,\dots,  [i_{lmn},j_{lmn},k_{lmn}])$, and
perform the subsequent greedy algorithm:     

\begin{algorithm}
For $s:=1$ to $lmn$ do
\begin{enumerate}
\item Set $x_{i_s,j_s,k_s}:= \min \{a^1_s, a^2_s, a^3_s\}$
\item $a^1_s := a^1_s -x_{i_s,j_s,k_s}, \, a^2_s := a^2_s -x_{i_s,j_s,k_s}, \, a^3_s := a^3_s -x_{i_s,j_s,k_s}$
 \end{enumerate}
\end{algorithm}
                                                            
Again, given a sequence $S$ of triples of indices, this greedy
algorithm maximizes each variable of $S$ in turn. When the algorithm
ends it will give always a feasible solution, in fact an integer
solution when the margins are integer. In \cite{rudolf} we have a 
necessary and sufficient condition on $S$ and
the cost matrix $c$ which guarantees that the solution is in fact an optimal
LP solution for costs $c_{i,j,k}$ associated to each entry:

\begin{lemma}  The generalized north-west rule algorithm finds a feasible
solution for the three-dimensional axial transportation problem for all
right-hand-side vectors $a_1$, $a_2$, and $a_3$ whose sum of entries
are equal. The solution is integer if the vectors $a_1,a_2,a_3$ are
integer. Moreover if there is a cost matrix cost matrix $c_{i,j,k}$,
which is a three-dimensional Monge sequence in the sense of
\cite{rudolf}, then the solution found is an optimal linear
programming solution for the minimization problem.
\end{lemma}

In the rest of this section, we will briefly introduce the notion of Gr\"obner bases
relevant for our project problems. Recall a {\em polynomial ideal} $I$ is a set of polynomials in $R={\mathbb Q}[x_1,\dots,x_n]$ that satisfies two properties: (1)
If $f,g$ are in $I$ then $f+g \in I$ (2) If $f \in I$ and $h \in R$ then
$fh$ is in $I$.  a Gr\"obner basis of an ideal $I$ is a special finite generating set for $I$ with special computational properties.
Their computational powers include the ability to answer membership questions for the ideal, computing intersections of ideals, computing projections of ideals, etc. Gr\"obner bases in general can be computed with the well-known \emph{Buchberger algorithm}. We are only interested in special kinds of ideals, called \emph{toric ideals} whose Gr\"obner bases are better behaved:
Given a matrix $A$ with integer entries, the toric ideal $I_A$ is the
ideal generated by the binomials of the form $x^u-x^v$ such that
$A(u-v)=0$.  Gr\"obner bases of toric ideals  have been explored
in the literature (see \cite{GB+CP, survey1,coxetal2} and references
therein). If we find a Gr\"obner basis $G_A=\{ x^{u_1} -x ^{v_1},
x^{u_2}-x^{v_2},\dots, x^{u_k}-x^{v_k}\}$ for $I_A$, it is well known
that the vectors $\Gamma = \{ u_1-v_1,u_2-v_2,\dots, u_k-v_k\}$ will have the
following properties. Let $P = \{x \, : \, Ax =b, \,\, x\geq 0\}$ be any polytope
that could be defined by the matrix $A$ and by a choice of an integral right-hand-side  vector $b$. If we form a graph
whose vertices are the lattice points of $P$ and we connect any pair
$x_1,x_2$ of them by an edge if there is a vector $u-v$ in $\Gamma$ such
that $x_1-u+v=x_2$ with $x_1 - u \geq 0$, 
then the resulting graph is connected \cite[Chapter 4]{GB+CP}.
Moreover if we orient the edges of this graph according to a
term order we used to compute the Gr\"obner basis above (where
the tail of an edge is bigger than its head)  this
directed graph will have a unique sink. Thus from any lattice point of
$P$ there is an``augmenting'' path to this sink. We will call
the process of traversing such an augmenting path a {\em reduction}. 
Moreover, we will refer to the elements in $\Gamma$ as {\em moves}.
It has been shown in \cite{HS1} that while toric ideals of general transportation polytopes are as nasty as those for general polytopes, axial transportation polytopes enjoy a rich decomposable structure that essentially allow us to build  Gr\"obner bases from the Gr\"obner bases of their slices. 
For instance, for 2-way tables we know everything about the Gr\"obner bases of their toric ideals 

\begin{theorem} \label{thm:circuits} 
Let $A$ be the $0/1$-matrix which is the matrix of the linear transformation
that computes the row and column sums of a given $2$-way table. Then $A$
is (totally) unimodular and hence its minimal universal Gr\"obner basis
consists of its circuits. These circuits are $2$-way tables whose row and 
column sums are zero and with entries in $\{0, +1, -1\}$ of minimal support.
\end{theorem}

For axial transportation polytope of size $m \times n
 \times k$, can we find a similarly nice Gr\"obner basis for some term 
 order. We explain several ways next.

\section{Integer Feasibility Testing is a Game on Tables} \label{algorithm}

We have seen that from Theorem \ref{theorem:main} the (integer) tables with specified margins  in the construction represent all the lattice solutions of the original IFP. Those will be the 
states of the game. In order to check the integer feasibility of $P$ we need to have a Gr\"obner basis (test set) of the axial transportation polyhedron $Q$ such that  the normal form of the  North-West-corner rule integer initial solution $v$ is a feasible solution of $P$ (if such a solution exists). Of course, such a
Gr\"obner basis exists right away: a Gr\"obner basis of the axial transportation
arrays with respect to an {\em elimination} term order where all
variables corresponding to the "forbidden" entries of the arrays are
bigger than the "enabled" entries will do the job. In principle, we are done. 
However in the rest of the section we explain how to find a more efficient solution
avoiding to use Buchberger algorithm. 

We construct such a Gr\"obner basis building from the Gr\"obner bases of $2$-way transportation problems slices of $3$-way tables (see Theorem \ref{thm:circuits} and discussion there). Moreover, we will
prove that we do not need to explicitly compute and store this
Gr\"obner basis in advance (which is a very large set of vectors). It
is enough to compute an element of the Gr\"obner basis "on the fly"
that will improve the current feasible solution. For our construction
we will follow the ideas presented in \cite{HS1}. There a similar construction was given
for any {\em decomposable} statistical problem. Fortunately, the $3$-way axial
problems are decomposable, so we can use those techniques. But we will
do this in a slightly more general way.  The first set of moves that
will make up a Gr\"obner basis is obtained as follows. Let $T$ be, for a 
fixed index value $k$, the 2-way transportation problem defined as $\{a_{i,j,k} \geq 0, \forall i,j,k,\, \sum_{j} a_{i,j,k} = u_{i,k},\, \forall (i,k), \, \sum_{i} a_{i,j,k} = v_{j,k}, \forall (j,k). \}$

Let $G_{\succ_1}, \ldots, G_{\succ_n}$ be $n$ different Gr\"obner bases 
of  $2$-way $l \times m$ 
transportation problems. Note that if the $l \times m$ table
$X$ with entries $X[i,j]$ is a Gr\"obner basis element, then
for each fixed $k$ the $3$-way table $Y$ with entries
$Y[i,j,k] = X[i,j]$ and $Y[i,j,t]=0$ for $t \neq k$ is a valid move for the transportation problem
$T$. Now let $\mathcal{F}(G_{\succ_1}, \ldots,  G_{\succ_n})$ be the set of moves obtained
this way from all elements in $G_{\succ_k}$ for all values of $k$. 
The following theorem is a modification  of Theorem 4.13 in \cite{HS1}, we omit its proof here:

\begin{theorem} \label{thm:nicemoves}
Let $G_{\succ_1}, \ldots , G_{\succ_n}$ be $n$ Gr\"obner bases of the $2$-way $l \times m$
transportation problem. Let $\succ'$ be the term order 
on the entries of the $l \times m \times n$ axial transportation problem
where $\{Y[i,j,1]\} \succ' \{Y[i,j,2]\} \succ' \cdots \succ' \{Y[i,j,n]\}$
and the entries in the $k$th horizontal slice $\{Y[i,j,k]\}$ are ordered with
respect to the term order $\succ_k$. Then $\mathcal{F}(G_{\succ_1}, \ldots,  G_{\succ_n})$
is a Gr\"obner basis with respect to $\succ'$. 
\end{theorem}

There is a second set of moves for the original transportation problem
$Q$ obtained from 2-way transportation problems. Now we will
describe these moves. This time let $T_{x,z}$ be the $2$-way planar
transportation problem. Let $G_{x,z}$ be a Gr\"obner basis
for this problem. Suppose $X_1 - X_2$ is an element in $G_{x,z}$
where $X_1$ and $X_2$ are nonnegative tables with entries
$X_1[i,k]$ and $X_2[i,k]$. We can ``lift'' such an element
to a move for the original problem $Q$ as follows. First note that
$X_1-X_2$ is homogeneous, i.e., $\sum_{i,k} X_1[i,k] = \sum_{i,k} X_2[i,k] = t$.
Second because we are in the setting of a $2$-way transportation problem,
for each $k$ we have $\sum_{i} X_1[i,k] = \sum_{i} X_2[i,k]$. Hence we can represent 
$X_1 - X_2$ as
\[
([i_1,k_1], [i_2,k_2], \ldots, [i_t, k_t]) - ([i_1',k_1], [i_2',k_2], \ldots, [i_t', k_t]). 
\] 
Here we allow that some indices $[i_s, k_s]$ repeated if the corresponding entry $X_1[\cdot, \cdot]$ 
(or $X_2[\cdot, \cdot]$) in the table is bigger than one. Now given 
a sequence of indices (again repetitions are allowed) $j_1, \ldots, j_t$ we get a move $Y_1 - Y_2$ for the transportation problem $Q$: $([i_1,j_1, k_1], [i_2,j_2, k_2], \ldots, [i_t, j_t, k_t]) - ([i_1',j_1, k_1], [i_2',j_2, k_2], \ldots, [i_t', j_t, k_t]).$

We let $\mathcal{L}(G_{x,z})$ to be the set of all moves obtained from all Gr\"obner basis elements in $G_{x,z}$ using all possible liftings. Similarly we can define $\mathcal{L}(G_{y,z})$. Now we claim we can put together 
$\mathcal{F}(G_{\succ_1}, \ldots,  G_{\succ_n})$, $\mathcal{L}(G_{x,z})$,
and $\mathcal{L}(G_{y,z})$ to get a Gr\"obner basis for the toric ideal of $l \times m \times n$ $3$-way
axial transportation problem. However, we first need to describe the appropriate 
term order. 

Given a $3$-way table $X$ we can compute ``projection'' tables (marginals) in any axial
direction. For instance $Proj_{x,z}(X)$ would be the $2$-way table whose
$(i,k)$ entry is $\sum_{j} X[i,j,k]$. 

\begin{lemma} \label{lem:termorder} Let $\succ^1$  and $\succ^2$ be term orders for 
$l \times n$ and  $m \times n$ planar tables and let  $\succ_1, \ldots, \succ_n$ 
be $n$ term orders forthe $l \times m$ planar tables. 
If $\succ'$ is the term order for $l \times m \times n$ $3$-way tables 
described in Theorem \ref{thm:nicemoves},  then the relation $\succ_*$ on
such $3$-way tables given by X $\succ_*$ X' if 
\begin{align*}
    Proj_{x,z}(X) & \succ^1 Proj_{x,z}(X') \, \mbox{or}\\
    Proj_{x,z}(X) = Proj_{x,z}(X') & \mbox{ and } Proj_{y,z}(X) \succ^2 Proj_{y,z}(X') \, \mbox{or} \\
    Proj_{x,z}(X) = Proj_{x,z}(X') & \mbox{ and } Proj_{y,z}(X) =  Proj_{y,z}(X') \mbox{ and } \\
                                   & X \succ' X'
\end{align*}
    
is a term order.
\end{lemma}

The proof of Lemma \ref{lem:termorder} is in the Appendix.

\begin{theorem} \label{thm:gb-construction}
Let $G_{x,z}$ and  $G_{y,z}$ be  Gr\"obner bases for the $l \times n$ and
$m \times n$ planar transportation problems with respect to 
the term orders $\succ^1$ and $\succ^2$, respectively. 
Also let  $G_{\succ_1}, \ldots, G_{\succ_n}$ be $n$ Gr\"obner bases 
for $ l \times m$ planar transportation problems with respect to the term orders
$\succ_1, \ldots, \succ_n$. Let $\succ'$ be 
the term order for the $l \times m \times n$ tables    
given in Theorem \ref{thm:nicemoves}. Then
the set
\[
G \quad = \quad \mathcal{L}(G_{x,z}) \cup \mathcal{L}(G_{y,z}) \cup \mathcal{F}(G_{\succ_1}, \ldots, G_{\succ_n}) 
\]
is a Gr\"obner basis for the $3$-way axial transportation problems with respect
to the term order $\succ_*$ of Lemma \ref{lem:termorder}.
\end{theorem}

 
Now we are ready to construct a Gr\"obner basis for $3$-way axial transportation problems 
with which we will solve the integer feasibility problem for any polytope $P$ after it has been
encoded as a $3$-way axial transportation polytope $Q$. In order to do this we will describe 
term orders $\succ^1$, $\succ^2$ and $\succ_1, \ldots, \succ_n$ and then use Theorem \ref{thm:gb-construction}. 
Recall that by the construction of $Q$, the tables corresponding to the feasible
solutions of $P$ will be a face $F$ given by forbidden entries which need to be set to zero. 
If $X$ is such a table where all the enabled entries are positive, then we get forbidden entries for 
tables $Proj_{x,z}(X)$, $Proj_{y,z}(X)$ which are forced to be equal to zero. 
Also each horizontal slice $X_1, \ldots,  X_n$ will have its own forbidden entries.
Let $F_{x,z}$, $F_{y,z}$ and $F_1, \ldots, F_n$ be these forbidden entries, 
and let $E_{x,z}$, $E_{y,z}$, 
and $E_1, \ldots, E_n$ be their complements, namely the enabled entries. We let
$w_{x,z}$ be the weight vector where $w_{x,z}(i,k) = 1$ if $(i,k) \in 
F_{x,z}$ and $w_{x,z}(i,k)=0$ if $(i,k) \in E_{x,z}$. We define 
$w_{y,z}$ and $w_1, \ldots, w_n$ in a similar way. We define the term order 
$\succ^1$ to be any elimination term order where $F_{x,z} \succ^1 E_{x,z}$ 
(and the entries $F_{x,z}$ and $E_{x,z}$
within themselves are ordered in an arbitrary but fixed way)
which refines the ordering giving by $w_{x,z}$. In other words, 
given two $l \times m \times n$ tables $X$ and $Y$, if
the weight of $Proj_{x,z}(X)$ is bigger than the weight of
$Proj_{x,z}(Y)$ then we declare $Proj_{x,z}(X) \succ^1 
Proj_{x,z}(Y)$. In the case of equality, we resort to the elimination term
order that breaks the tie.
Note that if the support of $Proj_{x,z}(Y)$ is
contained in $E_{x,z}$ and that of $P_{x,z}(X)$ is not, then
we immediately declare $Proj_{x,z}(X) \succ^1 Proj_{x,z}(Y)$.
We define $\succ^2$ and $\succ_1, \ldots, \succ_n$ similarly
in which the forbidden entries are eliminated. 

\begin{algorithm}
\caption{Integer Feasibility Testing Game}
\label{algo}
\smallskip \par
\noindent {\em Input:} A rational polytope  $P\subset\R^d$ presented in its representation  $P=\{x : Ax=b, \ x \geq 0\}$.
                                                                                
\noindent {\em Output:} YES or  NO depending on whether $P$ contains  an integer lattice
point.

\begin{enumerate}[\ 1.]
\item Compute the encoding of $P$ as a face of a $3$-way axial transportation
polytope $Q$. 
\item Use the North-West-corner rule to find an initial table $V$ that is a feasible integer solution in $Q$.
\item Use Gr\"obner basis elements with respect to $\succ_*$ constructed above 
to get the unique sink $W$. If the weight of $Proj_{x,z}(W)$ or of $Proj_{y,z}(W)$ is 
not zero, output NO.
\item Otherwise, using the Gr\"obner basis elements in $G_{x,z}$ and $G_{y,z}$ 
of weight zero (and their liftings) generate a set $\mathcal{S}$ of tables such that 
$\{Proj_{x,z}(T) \, : \, T \in \mathcal{S}\}$ and  $\{Proj_{y,z}(T) \, : \, T \in \mathcal{S}\}$ 
are the set of $2$-way $l \times n$ and $m \times n$ tables with the same row and column sums
as $Proj_{x,z}(W)$ and $Proj_{y,z}(W)$, and with their support in $E_{x,z}$ and $E_{y,z}$
respectively.
\item For each $T \in \mathcal{S}$ reduce $T$  just using the Gr\"obner basis elements in
$\mathcal{F}(G_{\succ_1}, \ldots, G_{\succ_n})$ to obtain $X$. If the weight 
of $X$ with respect to $w_k$ is zero for all $k=1, \ldots, n$, output YES. 
If no $T$ in $\mathcal{S}$ gives rise to such an $X$, output NO.
\end{enumerate}
\end{algorithm}

\begin{theorem} \label{theorem:alg}
Algorithm \ref{algo} solves the integer feasibility of any rational polytope $P$ by first transforming it into an 3-way axial transportation polytope $Q$ with specific 1-margins and finding a sequence of Gr\"obner basis moves from an initial 3-way array in $Q$ to another 3-way array in $Q$ with zeros in specified entries if one exists (equivalent to finding an integer solution for $P$).
\end{theorem}

\section{Learning to Play Games on 2-way Tables} \label{2waytables}


As a proof of concept, we now describe our method of playing table games using reinforcement learning on the polyhedral Gr\"obner bases systems we presented. Here we only consider the simpler special case of 2-way tables because we have an explicit list of all Gr\"obner basis moves already (See Theorem \ref{thm:circuits}). Similar ideas can be extended to 3-way axial transportation polytopes. In this family, the state space $\stateSpace$ consists of $m\times n$ tables of a 2-way transportation polytope with fixed 1-margins,  
\begin{equation*}
    \stateSpace = \{\state \in \mathbb{Z}^{m\times n} \cbar \sum_i \state_{i, j} = x_i,\, \sum_j \state_{i, j} = y_j,\, \state \ge 0 \}.
\end{equation*}

The action space is the \Grobner basis generators that connect any two $m \times n$ tables in $\stateSpace$. They are computed using Theorem \ref{thm:circuits}. It is noteworthy that the action space can be defined with more flexibility because one can consider the minimum set of actions that connects all tables, or with more complicated moves that are linear combinations of the minimum moves. There is a trade-off between the action space and the efficiency of solving the game. We refine our action space $\actionSpace$ as follows.
\begin{equation*}
    \actionSpace = \{\action \in \{-1, 0, 1\}^{m\times n} \cbar \sum_i \action_{i, j} = 0, \sum_j \action_{i, j} = 0\}.
\end{equation*}

Note that any element of the action space $\actionSpace$ is generated as integer combination of the  Gr\"obner basis of Theorem \ref{thm:circuits}, which suggests that for any two states $\state_1,\state_2 \in \stateSpace$, there exists a path connecting $\state_1, \state_2$ using moves from $\actionSpace$. In the most general form, our table games can be formulated as follows. 
\begin{align*}
\max_{\pi} \quad & \sum_t \reward_t(\state_t, \action_t)|_{\pi(\state_t) = \action_t} \\
& \text{s.t.} \quad \action_t \in \Omega(\state_t)\\
                  &  \quad\quad \state_{t+1} = \state_{t} + \action_t.
\end{align*}
where $\Omega(\state)$ is the set of valid moves for (table) $\state$:
    $\Omega(\state) = \{\action \cbar \action \in \actionSpace,\, \state + \action \ge \bzero\}.$

We pre-defined the terminating state of the game by specifying a set of entries $\goalEntries$ such that $\state_{(i, j)} = 0,\, \forall (i, j) \in \goalEntries$ as the terminating condition which mimics the forbidden entries of the axial transportation polytope computed in Theorem \ref{theorem:alg}. 
The reward function penalizes every step when the goal state is not reached:
\begin{align}
    \reward_t = 
    \begin{cases}
        \frac{1}{\norm{|s|}_{\infty}}, & \text{if}\quad \exists\, (i, j) \in \goalEntries,\, \state_{(i, j)} \ne 0, \\
        0, & \text{otherwise.}
    \end{cases}
\end{align}
The reward function encourages a policy $\pi$ that can reach a terminating state with the minimum number of steps.

\subsection{Reinforcement Learning Framework}
Our RL framework is based on Twin Delayed DDPG (TD3~\cite{TD3, DDPG}). TD3 is an off-policy method which requires a replay buffer $\RB$ to collect a set of transitions $(\state, \action, \state', \reward)$. There are two critic networks $\criticOne$ and $\criticTwo$ and an actor network $\actor$. Both critic networks are trained by minimizing the mean squared Bellman error: 
\begin{equation*}
    L_Q(\theta_i) = \E_{(\state, \action, \state', \reward)\sim \RB} \norm{y(\state', \reward) - \criticAny(\state, \action)}^2, \, i=1,2.
\end{equation*}
The target $y$ is defined as the minimum of two fixed target critic networks to prevent overestimation of the $Q$ function. 
\begin{equation*}
    y(\state', \reward) = \reward + \gamma \min_{i=1, 2} \criticTarget (\state', \actorTarget(\state')).
\end{equation*}

\subsubsection{Structured Action Prediction}

\begin{figure}
    \centering
    \includegraphics[width=\linewidth]{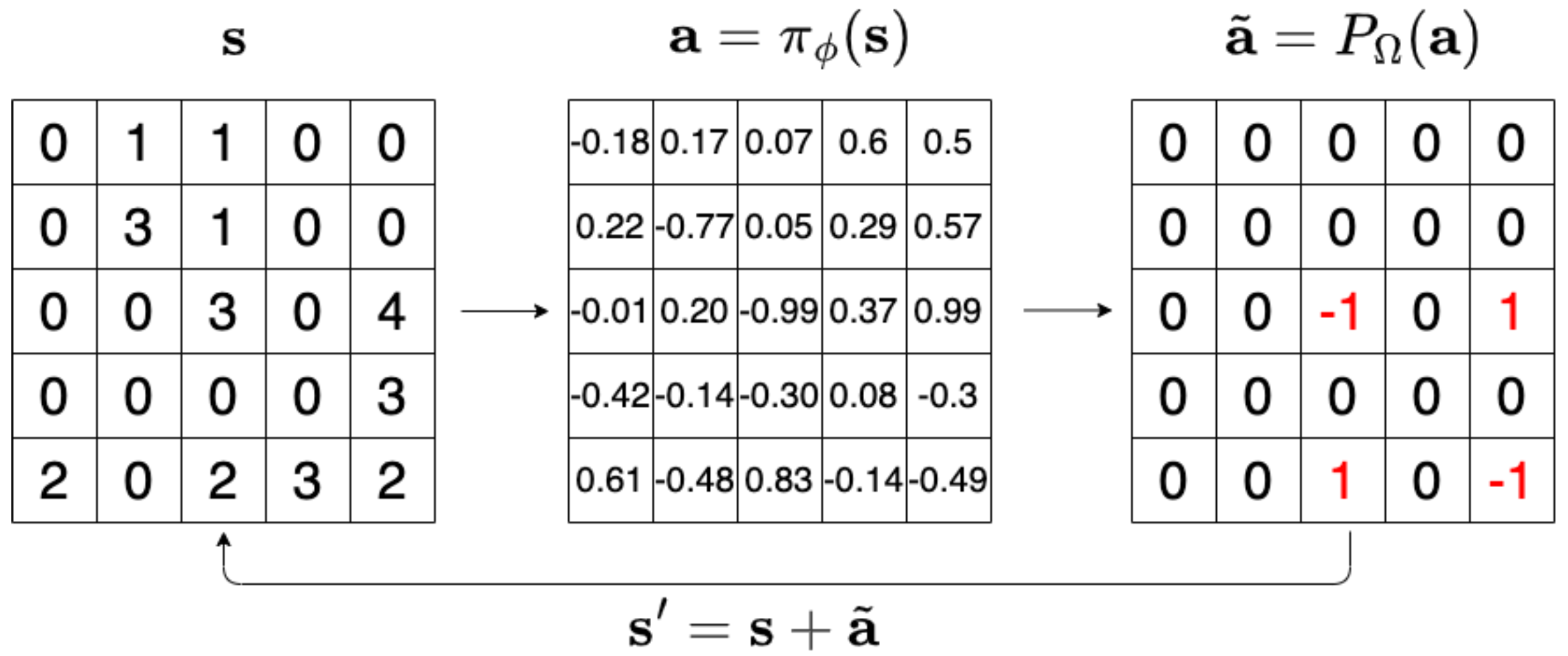}
    \caption{The interaction between the actor and the environment. The actor network $\actor$ predicts a continuous move $\action$, which is then projected to a discrete $\actionProj$ in the set $\Omega(\state)$ before being applied to the environment.}
    \label{fig:action_proj}
\end{figure}

Due to the fact that our action space is discrete, one natural way is to predict the action as a classification problem at each time step. However, its feasibility is hindered by the large number of actions to compute and store in advance. We instead make the actor network predict a continuous action $\action = \actor(\state)$. Then, we obtain the integral solution $\actionProj = P_{\Omega(\state)}(\action)$ by projecting $\action$ to the feasible set of discrete actions encoded by the following integer program.
\begin{align}
\min_{\actionPos, \actionNeg} \quad & \sum_{i, j} ||\actionPos - \actionNeg - \action||_d \nonumber \\
\text{s.t.} \quad & \sum_{i} \actionPos_{i, j} - \actionNeg_{i, j} = \bzero,\, \sum_{j} \actionPos_{i, j} - \actionNeg_{i, j} = \bzero \nonumber \\
                & \quad \actionPos + \actionNeg \le 1 ,\, \sum_{i, j} \actionPos_{i, j} \ge 1, \label{eq:action_non_zero}\\
                  & \quad \actionPos,\, \actionNeg \in \{0, 1\}^{m\times n},\nonumber
\end{align}
and we obtain the projected discrete action by $\actionProj = \actionPos - \actionNeg$. We add constraint~\eqref{eq:action_non_zero} to exclude the action with all zeros. An illustration of the process is shown in Figure~\ref{fig:action_proj}. Splitting $\actionProj$ into two binary variables $\actionPos$ and $\actionNeg$ not only speeds up the projection but also provides more flexibility on the constraints of the actions. For instance, we can bound the number of non-zero elements in the action by adding $c_1 \le \sum_{i, j} \actionPos_{i, j} \le c_2$ to the system. 

To train the critic network, we consider the projection operator a deterministic part of the environment and directly learn $Q(\state, \action)$. Similar strategies can be found in \citet{RL_large_action_space}. It is worthwhile to mention that one can also learn $Q(\state, \actionProj)$. The non-differentiable projection layer can be tackled by straight-through estimators~\cite{straight_through_estimator,vqvae}. However, for every mini-batch update, calculating a target in the temporal difference learning requires the projection operations to obtain $\actionProj$. The projection will become a bottleneck and significantly slows down the whole training process. We also observed in experiments that learning $Q(\state, \actionProj)$ has no obvious performance gain. The actor network is updated for each mini-batch $B$ by gradient ascent 
\begin{equation}
    \nabla J(\phi) = \frac{1}{|B|} \sum \nabla_{\action} \critic(\state, \action) \nabla_{\phi} \actor(\state). \label{eq:actor_gradient}
\end{equation}


\subsection{Learning from Demonstrations}
Learning from demonstrations is an effective strategy to improve sample efficiency. Due the the large action space of our problem, we generate a set of demonstrations $\mathcal{B}_D$ using a greedy algorithm to speed up the learning. We make sure that a fixed portion of samples in the mini-batch are drawn from the demonstration during each training step. Since the demonstrations generated by the greedy algorithm are not perfect, we adopt the Q-filter strategy~\cite{Q_filter} to only enforce a supervised learning loss when the demonstrated action $\action$ has a higher $Q$ value than the actor's action. Thus, the demonstration loss on mini-batch $B_D$ using demonstrated action $\action_D$ can be summarized as:
\begin{equation*}
    L_D = \frac{1}{|B_D|} \sum ||(\action_D - \actor(\state))||^2 \cdot \mathbbm{1}_{\critic(\state, \action_D) > \critic(\state, \actor(\state))}.
\end{equation*}



\begin{figure*}[t]
    \captionsetup[subfloat]{labelformat=empty}
    \centering
    \subfloat{\includegraphics[width=0.3\linewidth]{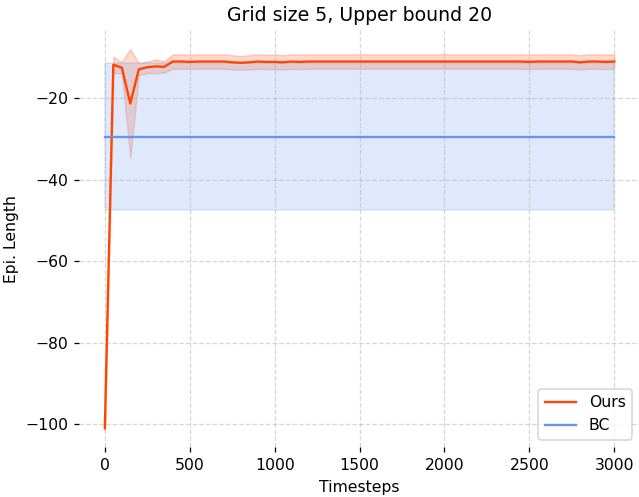}}
    \hfil
    \subfloat{\includegraphics[width=0.3\linewidth]{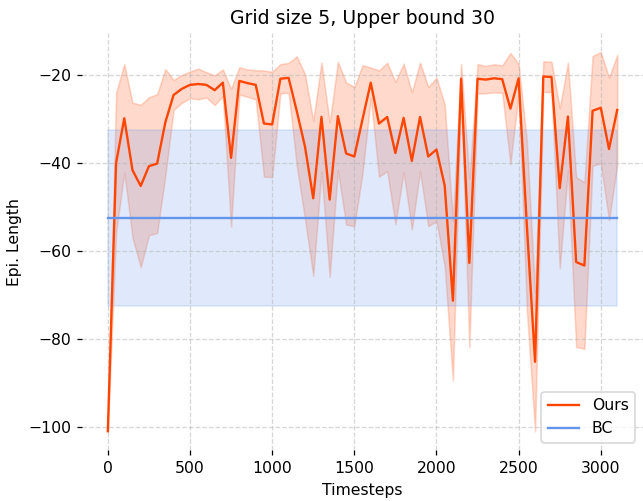}}
    \hfil
    \subfloat{\includegraphics[width=0.3\linewidth]{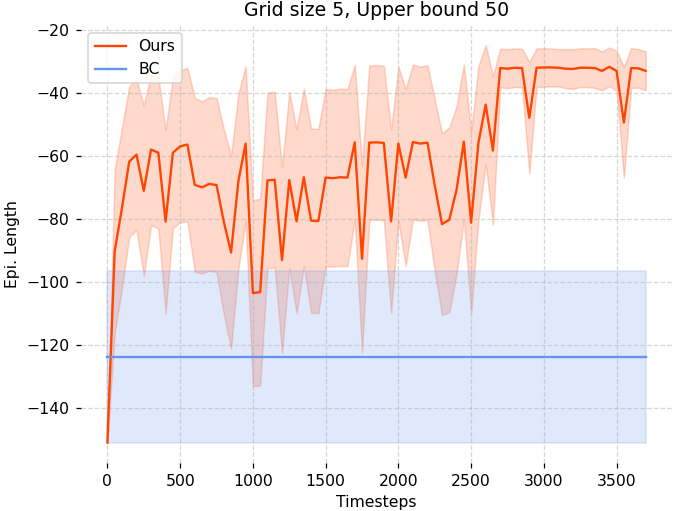}}
    \vspace{-0ex}
    \label{fig:results_gs5}
    \caption{The learning curve of our RL agent compared with the behavior cloning baseline in $5\times 5$ table games.  }
\end{figure*}

\begin{figure*}[t]
    \captionsetup[subfloat]{labelformat=empty}
    \centering
    \subfloat{\includegraphics[width=0.3\linewidth]{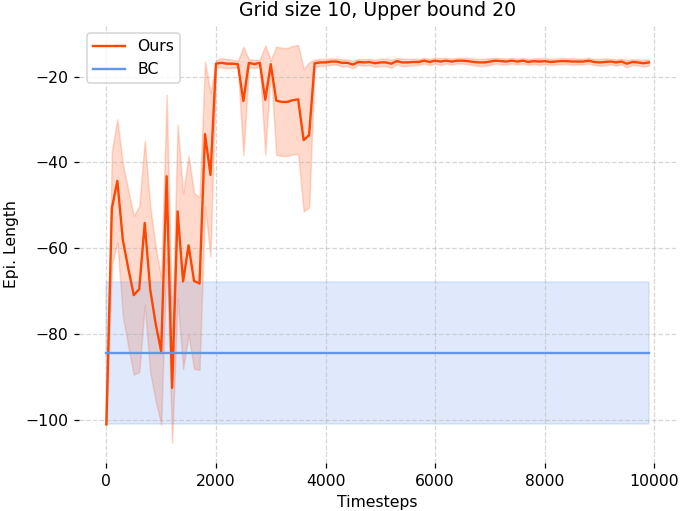}}
    \hfil
    \subfloat{\includegraphics[width=0.3\linewidth]{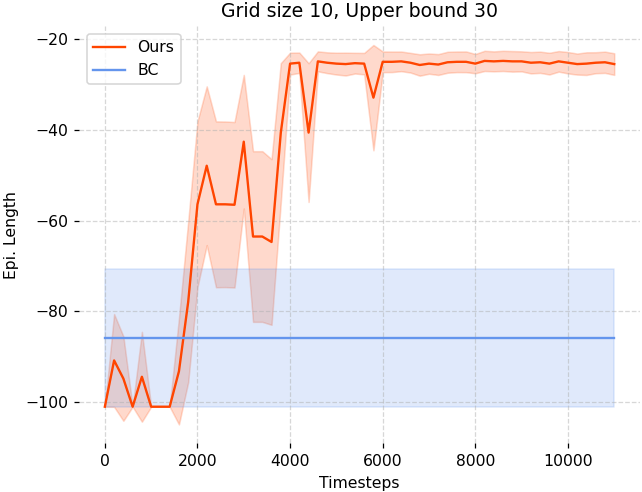}}
    \hfil
    \subfloat{\includegraphics[width=0.3\linewidth]{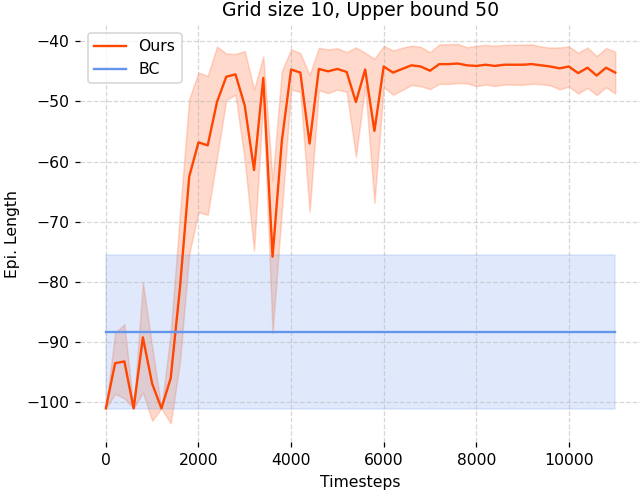}}
    \vspace{-0ex}
    \label{fig:results_gs10}
    \caption{The learning curve of the RL agent compared to BC model on $10 \times 10$ tables.}
\end{figure*}

\section{Experiments}
\subsection{Data Collection}
The goal entries in $\goal$ are selected randomly and remain fixed. The initial state is also randomly generated, and we randomly generate forbidden entries and fill in other entries with random numbers. To start an episode, we randomly initialize the starting table $\state$. To improve the training stability, we add lower and upper bounds on the 1-margins, i.e., $LB \le \sum_i\state_i \le UB$. As a result, the 1-margins are fully specified by the initial state and remain the same throughout the episode. The moves of the demonstrating greedy algorithm are calculated state-by-state by selecting a valid move that maximizes the values of the entries in $\goalEntries$.

\subsection{Network Architecture and Baseline}
The actor and critics are parameterized by convolutional neural networks. networks consist of multiple convolutional blocks where each block has a convolution layer followed by batch normalization~(BN) and ReLU. The number of blocks depends on the size of the board. All convolution layers maintain the same spatial dimensions. The final block of the actor network has a convolution layer followed by tanh. In the critic network, we apply a global average pooling layer before the final linear layer to predict the $Q$ value. 

We also train a baseline model on the demonstration set using behavior cloning (BC) \cite{dagger_bc, BC_survey}. The BC model is also a neural network with the same architecture as the actor network. 

\subsection{Training Details}
We collect 100 demonstrations prior to the RL training and we sample $10\%$ of transitions from the demonstration buffer for every mini-batch update. We use Adam~\cite{Adam} as the optimizer with a learning rate $10^{-4}$ for both actor and critics. The discount factor $\gamma$ is $0.99$. We use a batch size of $32$. The exploration of DDPG-styled algorithm is achieved by adding a Gaussian noise to the actor $\action = \actor(\state) + \epsilon.$ We use $\epsilon \sim \mathcal{N}(0, 0.2)$ throughout the experiment. 

\begin{figure}[h]
    \centering
    \includegraphics[width=0.7\linewidth]{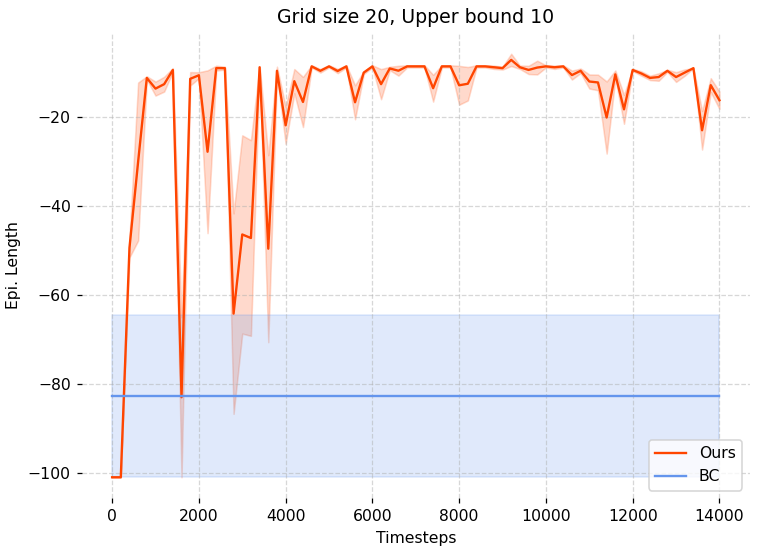}
    \caption{The learning curve of the RL agent compared to BC model on $20 \times 20$ tables.}
    \label{fig:results_gs20}
\end{figure}

\subsection{Results}
To demonstrate the ability of our method to learn complex moves, we test our agent on 2-way table of various grid (table) sizes with increasing maximum elements. Larger maximum entries in the table would lead to longer horizons. The results on $5\times 5$ table games is shown in Figure 2. The mean negative episode length is shown in as the curve and the shaded region denotes the half standard deviation. The BC model can achieve relatively good performance but deteriorates as the upper bound on the entry increases. The RL model outperforms the BC model by a margin and the advantage becomes more significant with larger upper bounds. In Figure~3 we observe similar performance on $10\times10$ games. We tested three 1-margin bounds $\{20, 30, 50\}$. The BC model does not perform well even with $UB = 20$. On the other hand, our RL agent is able to consistently reach the goal tables with relatively short paths. The results on $20\times 20$ tables are shown in Figure~\ref{fig:results_gs20}. The increase of the grid size also increases the action space exponentially. The RL agent can solve the problem whereas the BC model has a very low success rate and it takes longer steps to reach the goal state. 

\subsection{Generalization Test}
\begin{table}[h]
    \centering
    \begin{tabular}{c|cccccc}
       $UB$ & $20$ & $40$ & $60$ & $80$ & $100$ & $140$ \\
          \hline
       $GS=5$  & $1.0$ & $1.0$ & $0.90$ & $0.60$ & $0.72$ & $0.39$ \\
       $GS=10$  & $1.0$ & $1.0$ & $0.79$ & $0.49$ & $0.34$ & $0.32$ \\
    \end{tabular}
    \caption{We test the trained model with margin bound $20$ on increased margins and compare the how the success rate varies. The success rates drop as we increase the bound, but the model is able to maintain a decent success rate which shows its generalization ability.}
    \label{tab:ub_increase}
\end{table}
In this section we wonder if the model can solve games that are unseen in the training data. We set up the experiment by training the model on the fixed upper 1-margin bound $20$ with grid size $5$ and $10$, and test its success rates on larger bounds with the same corresponding grid size. The results in Table~\ref{tab:ub_increase} demonstrate that the trained model can solve games that it never experiences during the policy training step. This further suggests that the model can successfully recognize patterns in the table and produces corresponding moves regardless of their magnitudes. 

\section{Conclusions and Future Work}

We proposed an algorithm that converts the integer feasibility problem into a game on tables. We formulated the table game as a reinforcement learning problem and developed novel techniques to tackle the algebraic structure of the action space. The experimental results on 2-way tables show the potential of solving integer feasibility problems using the \Grobner bases approach. Training on 3-way tables is the next stage of our work. Since IFP is an NP-complete problem, in some cases our game algorithm may have to visit all (exponentially many) tables before concluding there is no solution. It is an open problem to explore ways to cut the search as our games have no solution sometimes.



\bibliography{aaai23}

\pagebreak

\onecolumn
\section{Appendix}

In this appendix we include technical proofs of some of the main theorems.
Including an explanation of how any rational polyhedron can be
rewritten as a face of a 3-way axial transportation polytopes
with fixed 1-marginals.

\subsection{Proofs of Lemma \ref{lem:termorder} and Theorem \ref{theorem:alg}}


\begin{proof}[Proof. (Lemma \ref{lem:termorder})]
If $X \neq X'$ it is clear that either $X \succ_*  X'$ or $X' \succ_* X$, and the relation
is compatible with adding the same table $Y$ to $X$ and $X'$. So we just need to show that
$\succ^*$ is transitive. So let $X \succ_* X'$ and $X' \succ_* X''$. There is a total
of nine possibilities to be checked depending on how the tables are aligned, so we give one
of these for illustration. Suppose $X \succ_* X'$ because 
$Proj_{x,z}(X) = Proj_{x,z}(X')$ but $Proj_{y,z}(X) \succ^2 Proj_{y,z}(X')$, and also suppose
that $X' \succ_* X''$ because $Proj_{x,z}(X') \succ^1 Proj_{x,z}(X'')$. 
Then $Proj_{x,z}(X) = Proj_{x,z}(X') \succ^1 Proj_{x,z}(X'')$. Hence $X \succ_* X''$.
\end{proof}

\begin{proof} (of Theorem \ref{theorem:alg})
First we show that if $P$ is feasible then the $w_{x,z}$-weight of $Proj_{x,z}(W)$
and the $w_{y,z}$-weight of $Proj_{y,z}(W)$ must be zero. Suppose $U$ is a table
corresponding to a feasible solution of $P$. Then clearly the $w_{x,z}$-weight of 
$Proj_{x,z}(U)$ and the $w_{y,z}$-weight of $Proj_{y,z}(U)$ are zero. But then if one of these weights
for $W$ were bigger than zero we would get a contradiction to the assumption that
$W$ is the unique sink obtained by reduction of $V$ with respect to $\succ_*$.  
Now because the $w_{x,z}$-weight of both $Proj_{x,z}(W)$ and 
$Proj_{x,z}(U)$ are zero, and since $Proj_{x,z}(W)$ is the unique sink 
of $2$-way $l \times n$ tables with row and column sums equal 
to those of $Proj_{x,z}(V)$, there is a sequence of Gr\"obner basis
elements in $G_{x,z}$ which reduce $Proj_{x,z}(U)$ to 
$Proj_{x,z}(W)$. These elements must have $w_{x,z}$-weight zero.
This means that we can reverse these moves and use their liftings 
to obtain a table $W'$ such that $Proj_{x,z}(W') = Proj_{x,z}(U)$. 
Note that $Proj_{y,z}(W') = Proj_{y,z}(W)$, and we can use the
same argument above to conclude that one can reach to a table $T$
using lifted elements of $G_{y,z}$ with $w_{y,z}$-weight zero
such that $Proj_{y,z}(T)=Proj_{y,z}(U)$. Of course, we also
have $Proj_{x,z}(T) = Proj_{x,z}(U)$. The table $T$ will be
an element of $\mathcal{S}$ in Step 4 above,  if we use the (reversed)
elements of $G_{x,z}$ and $G_{y,z}$ of respective weights zero to
generate all possible $2$-way $l \times n$ and $m \times n$ tables 
with the same row and column sums as $Proj_{x,z}(W)$ and
$Proj_{y,z}(W)$. By this construction, for each $1 \leq k \leq n$ the
horizontal slices $T_k$ and $U_k$ have identical row and
column sums. Since the $w_k$-weight of $U_k$ is zero, if the same weight
of  $T_k$ is  not zero, one can find a sequence of Gr\"obner 
basis elements in $G_{\succ_k}$ (and hence in  $\mathcal{F}(G_{\succ_1}, \ldots, G_{\succ_n})$)
to obtain $X$ where the $w_k$-weight of $X_k$ is zero. 
Because $Proj_{x,z}(X)=Proj_{x,z}(T)=Proj_{x,z}(U)$ and
$Proj_{y,z}(X)=Proj_{y,z}(T)=Proj_{y,z}(U)$,  the table $X$ corresponds 
to a feasible solution of $P$. 
\end{proof}




\subsection{Preprocessing: Coefficient Reduction}
\label{Binary}

Let $P=\{y\geq 0:Ay=b\}$ where $A=(a_{i,j})$ is an integer matrix and
$b$ is an integer vector. We represent it as a polytope $Q=\{x\geq
0:Cx=d\}$, in polynomial-time, with a $\{-1,0,1,2\}$-valued matrix
$C=(c_{i,j})$ of coefficients, as follows. Consider any variable $y_j$
and let $k_j:=\max\{\lfloor \log_2 |a_{i,j}| \rfloor\,:\,i=1,\dots
m\}$ be the maximum number of bits in the binary representation of the
absolute value of any $a_{i,j}$. We introduce variables
$x_{j,0},\dots,x_{j,k_j}$, and relate them by the equations
$2x_{j,i}-x_{j,i+1}=0$. The representing injection $\sigma$ is defined
by $\sigma(j):=(j,0)$, embedding $y_j$ as $x_{j,0}$.  Consider any
term $a_{i,j}\,y_j$ of the original system.  Using the binary
expansion $|a_{i,j}|=\sum_{s=0}^{k_j}t_s 2^s$ with all
$t_s\in\{0,1\}$, we rewrite this term as $\pm\sum_{s=0}^{k_j}t_s
x_{j,s}$. It is easy to see that this procedure provides a new representation,
and we get the following.

\bl{Preprocessing}
Any rational polytope $P=\{y\geq 0:Ay=b\}$ is polynomial-time
representable as a polytope $Q=\{x\geq 0:Cx=d\}$ with
$\{-1,0,1,2\}$-valued defining matrix $C$.
\el

\subsection{Representing Polytopes as 3-way Transportation Polytopes with
1-marginals and forbidden entries}
\label{Plane}

Let $P=\{y\geq 0:Ay=b\}$ where $A=(a_{i,j})$ is an $m\times n$ integer matrix
and $b$ is an integer vector: we assume that $P$ is bounded and hence a
polytope, with an integer upper bound $U$ (which can be derived from the
Cramer's rule bound) on the value of any coordinate $y_j$ of any $y\in P$.

For each variable $y_j$, let $r_j$ be the largest between the sum of the
positive coefficients of $y_j$ over all equations and the sum of absolute
values of the negative coefficients of $y_j$ over all equations,
$$r_j\,:=\, \max
\left(\sum_k\{a_{k,j}:a_{k,j}>0\}\,,\,\sum_k\{|a_{k,j}|:a_{k,j}<0\}\right)\ .$$
Let $r:=\sum_{j=1}^n r_j$, $R:=\{1,\dots,r\}$, $h:=m+1$ and $H:=\{1,\dots,h\}$.
We now describe how to construct vectors $u,v\in\Z^r, w\in\Z^h$,
and a set $E\subset R\times R\times H$ of triples - the ``enabled",
non-``forbidden" entries - such that the polytope $P$ is represented as the
corresponding transportation polytope of $r\times r\times h$ arrays
with plane-sums $u,v,w$ and only entries indexed by $E$ enabled,
\begin{align*}
    T \, = \, \{ \, x\in\R_{\geq 0}^{r\times r\times h}\ \cbar
\ x_{i,j,k}=0\ \, \mbox{for all}\ \, (i,j,k)\notin E\,,\,
\ \, \mbox{and}\ \, 
 \sum_{i,j} x_{i,j,k}=w_k\,,
\ \sum_{i,k} x_{i,j,k}=v_j\,,\ \sum_{j,k} x_{i,j,k}=u_i\, \}\ .
\end{align*}

We also indicate the injection
$\sigma:\{1,\dots,n\}\longrightarrow R\times R\times H$
giving the desired embedding of the coordinates $y_j$ as the
coordinates $x_{i,j,k}$ and the representation of $P$ as $T$
(see paragraph following Theorem \ref{theorem:main}).

Basically, each equation $k=1,\dots,m$ will be encoded in a ``horizontal plane"
$R\times R\times\{k\}$ (the last plane $R\times R\times\{h\}$ is included
for consistency and its entries can be regarded as ``slacks"); and
each variable $y_j$, $j=1,\dots,n$ will be encoded in a ``vertical box"
$R_j\times R_j\times H$, where $R=\biguplus_{j=1}^n R_j$ is the natural
partition of $R$ with $|R_j|=r_j$, namely with $R_j:=\{1+\sum_{l<j}r_l,\dots,\sum_{l\leq j}r_l\}$.

Now, all ``vertical" plane-sums are set to the same value $U$, that is,
$u_j:=v_j:=U$ for $j=1,\dots,r$. All entries not in the union
$\biguplus_{j=1}^n R_j\times R_j\times H$
of the variable boxes will be forbidden. We now describe the enabled entries
in the boxes; for simplicity we discuss the box $R_1\times R_1\times H$,
the others being similar. We distinguish between the two cases
$r_1=1$ and $r_1\geq 2$. In the first case, $R_1=\{1\}$; the box, which
is just the single line $\{1\}\times\{1\}\times H$, will have exactly two
enabled entries $(1,1,k^+),(1,1,k^-)$ for suitable $k^+$, $k^-$ to be
defined later. We set $\sigma(1):=(1,1,k^+)$, namely embed $y_1=x_{1,1,k^+}$.
We define the {\em complement} of the variable $y_1$ to be
${\bar y}_1:=U-y_1$ (and likewise for the other variables). The vertical
sums $u,v$ then force ${\bar y}_1=U-y_1=U-x_{1,1,k^+}=x_{1,1,k^-}$, so the
complement of $y_1$ is also embedded. Next, consider the case $r_1\geq 2$.
For each $s=1,\dots, r_1$, the line $\{s\}\times \{s\}\times H$
(respectively, $\{s\}\times \{1+ (s \mod r_1)\}\times H$) will contain
one enabled entry $(s,s,k^+(s))$ (respectively, $(s,1+ (s \mod r_1),k^-(s))$.
All other entries of $R_1\times R_1\times H$ will be forbidden.
Again, we set $\sigma(1):=(1,1,k^+(1))$, namely embed $y_1=x_{1,1,k^+(1)}$;
it is then not hard to see that, again, the vertical sums $u,v$ force
$x_{s,s,k^+(s)}=x_{1,1,k^+(1)}=y_1$ and
$x_{s,1+ (s \small \mod r_1),k^-(s)}=U-x_{1,1,k^+(1)}={\bar y}_1$
for each $s=1,\dots, r_1$. Therefore, both $y_1$ and ${\bar y}_1$
are each embedded in $r_1$ distinct entries.

To clarify the above description it is helpful to visualize the
$R\times R$ matrix $(x_{i,j,+})$ whose entries are the vertical line-sums
$x_{i,j,+}:=\sum_{k=1}^h x_{i,j,k}$. 

Next we encode the equations by defining the horizontal
plane-sums $w$ and the indices $k^+(s), k^-(s)$ above as follows.
For $k=1,\dots,m$, consider the $k$th equation $\sum_j a_{k,j}y_j=b_k$.
Define the index sets $J^+:=\{j:a_{k,j}>0\}$ and $J^-:=\{j:a_{k,j}<0\}$,
and set $w_k:=b_k+U\cdot\sum_{j\in J^-}|a_{k,j}|$.
The last coordinate of $w$ is set for consistency
with $u,v$ to be $w_h=w_{m+1}:=r\cdot U-\sum_{k=1}^m w_k$.
Now, with ${\bar y}_j:=U-y_j$ the complement of
variable $y_j$ as above, the $k$-th equation can be rewritten as
\begin{align*}
\sum_{j\in J^+} a_{k,j}y_j+\sum_{j\in J^-} |a_{k,j}|{\bar y_j}
 = \sum_{j=1}^n a_{k,j}y_j+U\cdot \sum_{j\in J^-}|a_{k,j}| 
 = b_k+U\cdot\sum_{j\in J^-}|a_{k,j}| 
 = w_k.
\end{align*}

To encode this equation, we simply ``pull down" to the
corresponding $k$th horizontal plane as many copies of each variable
$y_j$ or ${\bar y}_j$ by suitably setting $k^+(s):=k$ or $k^-(s):=k$.
By the choice of $r_j$ there are sufficiently
many, possibly with a few redundant copies which are absorbed in
the last hyperplane by setting $k^+(s):=m+1$ or $k^-(s):=m+1$.
For instance, if $m=8$, the first variable $y_1$ has $r_1=3$ as above,
its coefficient $a_{4,1}=3$ in the fourth equation is positive,
its coefficient $a_{7,1}=-2$ in the seventh equation is negative,
and $a_{k,1}=0$ for $k\neq 4,7$, then we set $k^+(1)=k^+(2)=k^+(3):=4$
(so $\sigma(1):=(1,1,4)$ embedding $y_1$ as $x_{1,1,4}$),
$k^-(1)=k^-(2):=7$, and $k^-(3):=h=9$.

This way, all equations are suitably encoded, and Theorem \ref{theorem:main} follows from the construction outlined above and Lemma \ref{Preprocessing}.

\begin{customthm}{1}[Restated]
Any rational polytope $P=\{y\in\R_{\geq 0}^n:Ay=b\}$ is polynomial-time
representable as a plane-sum entry-forbidden $3$-way transportation polytope
\begin{align*}
    T \ =\ \{ \,  x\in\R_{\geq 0}^{r\times r\times h}\ \cbar
\ x_{i,j,k}=0\ \, \mbox{for all}\ \, (i,j,k)\notin E\,,\,
\ \, \mbox{and}\ \, 
 \sum_{i,j} x_{i,j,k}=w_k\,,
\ \sum_{i,k} x_{i,j,k}=v_j\,,\ \sum_{j,k} x_{i,j,k}=u_i\, \}\ .
\end{align*}
\end{customthm}



\end{document}

%% file: dfn.tex
\newtheorem{theorem}{Theorem}
\newtheorem{lemma}{Lemma}

\newtheorem{corollary}{Corollary}
\newtheorem{proposition}{Proposition}

\newtheorem{problem}{Problem}

\newtheorem{example}[theorem]{Example}{\bfseries}{\normalfont}
{\bfseries}{\normalfont}

\def\R{\mathbb{R}}
\def\Z{\mathbb{Z}}

\def \mod {\,\mbox{mod\,}}

\newcommand{\bt}[1]{\begin{theorem}\label{#1}}
\newcommand{\bc}[1]{\begin{corollary}\label{#1}}
\newcommand{\bl}[1]{\begin{lemma}\label{#1}}
\newcommand{\bp}[1]{\begin{proposition}\label{#1}}
\newcommand{\bpro}[1]{\begin{problem}\label{#1}}
\newcommand{\be}[1]{\begin{example}\rm\label{#1}}
\newcommand{\ba}[1]{\begin{algorithm}\rm\label{#1}}
\newcommand{\bd}[1]{\begin{definition}\rm\label{#1}}

\newcommand{\et}{\end{theorem}}
\newcommand{\ec}{\end{corollary}}
\newcommand{\el}{\end{lemma}}
\newcommand{\ep}{\end{proposition}}
\newcommand{\epro}{\end{problem}}
\newcommand{\ee}{\end{example}}
\newcommand{\ea}{\end{algorithm}}
\newcommand{\ed}{\end{definition}}

\newcommand{\reward}{\mathbf{r}}
\newcommand{\actionSpace}{\mathcal{A}}
\newcommand{\action}{\mathbf{a}}
\newcommand{\actionProj}{\Tilde{\mathbf{a}}}
\newcommand{\stateSpace}{\mathcal{S}}
\newcommand{\state}{\mathbf{s}}
\newcommand{\goal}{\mathbf{g}}
\newcommand{\goalEntries}{\mathcal{G}}
\newcommand{\Grobner}{\text{Gr\"{o}bner }}

\newcommand{\cbar}{\, | \,}
\newcommand{\actor}{\pi_\phi}
\newcommand{\actorTarget}{\pi_{\phi'}}
\newcommand{\criticTarget}{Q_{\theta'_i}}
\newcommand{\critic}{Q_\theta}
\newcommand{\criticOne}{Q_{\theta_{1}}}
\newcommand{\criticTwo}{Q_{\theta_{2}}}
\newcommand{\criticAny}{Q_{\theta_{i}}}
\newcommand{\actionPos}{\action^+}
\newcommand{\actionNeg}{\action^-}

\newcommand{\bzero}{\mathbf{0}}

\newcommand{\E}{\mathbb{E}}
\newcommand{\RB}{\mathcal{B}}
\newcommand{\norm}[1]{\left\lVert#1\right\rVert}

\newtheorem{innercustomgeneric}{\customgenericname}
\providecommand{\customgenericname}{}
\newcommand{\newcustomtheorem}[2]{%
  \newenvironment{#1}[1]
  {%
   \renewcommand\customgenericname{#2}%
   \renewcommand\theinnercustomgeneric{##1}%
   \innercustomgeneric
  }
  {\endinnercustomgeneric}
}

\newcustomtheorem{customthm}{Theorem}